\pdfoutput=1

\documentclass[11pt]{article}

\usepackage{acl}
\usepackage{times}
\usepackage{latexsym}

\usepackage[T1]{fontenc}

\usepackage[utf8]{inputenc}
\usepackage{inconsolata}

\usepackage{microtype}
\usepackage{times}
\usepackage{latexsym}
\usepackage{multirow}
\usepackage{url}

\usepackage{adjustbox}
\usepackage{graphicx}
\usepackage[scaled=.90]{helvet}
\usepackage{amsfonts}
\usepackage{float}
\restylefloat{table}
\usepackage{booktabs}
\usepackage{amsmath}
\usepackage{enumitem}
\usepackage{multirow}
\usepackage{subcaption}
\usepackage{cleveref}
\crefformat{section}{\S#2#1#3}
\crefformat{subsection}{\S#2#1#3}

\usepackage[normalem]{ulem}

\usepackage{caption}

\usepackage{xspace}
\newcommand{\ratsi}{RAT-SI\xspace}
\newcommand{\ratcat}{RAT-CAT\xspace}
\newcommand{\ratsep}{RAT-SEP\xspace}

\usepackage{todonotes}

\title{Improving Retrieval Augmented Neural Machine Translation by Controlling Source and Fuzzy-Match Interactions}

\author{Cuong Hoang\thanks{\hspace{1.5mm}Work done while the authors were at AWS AI Labs.}, Devendra Sachan\footnotemark[1], Prashant Mathur, 
Brian Thompson, Marcello Federico\\ 
AWS AI Labs \\
\href{mailto:pramathu@amazon.com}{pramathu@amazon.com}}

\date{}

\begin{document}
\maketitle
\begin{abstract}

We explore zero-shot adaptation,
where a general-domain model has access to  customer or domain specific parallel data 
at inference time, but not during training. 
We build on the idea of Retrieval Augmented Translation (RAT) where top-k in-domain fuzzy matches are found for the source sentence, and target-language translations of
those fuzzy-matched sentences are provided to the translation model at inference time. 
We propose a novel architecture to control interactions between 
a source sentence and the top-k fuzzy target-language matches, and compare it to  
architectures from prior work.  
We conduct experiments in two language pairs (En-De and En-Fr) by training models on WMT data and testing them with five and seven multi-domain datasets, respectively. Our approach consistently outperforms the alternative architectures, 
improving BLEU across language pair, domain, and number $k$ of fuzzy matches. 

\end{abstract}

\begin{figure*}[ht]
\centering
  \includegraphics[scale=0.31]{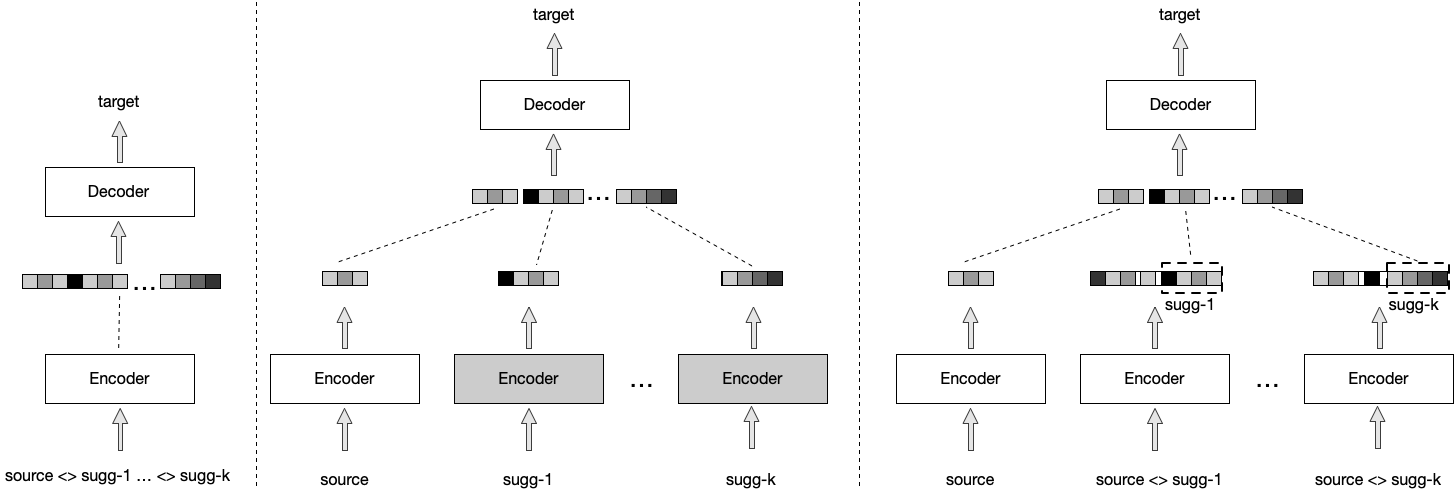}
  \caption{Architectures for retrieval augmented NMT. Left: Plain transformer ingesting source and retrieved fuzzy matches concatenated with a separator symbol \cite{bulte-tezcan-2019-neural}, denoted herein as \ratcat. Center: Transformer with dual encoder, one for encoding the source and one for encoding each retrieved fuzzy-matches, inspired by \citet{he-etal-2021-fast}, denoted herein as \ratsep. Right: Transformer separately encoding the source and each source + fuzzy-match pair (this work), denoted herein as \ratsi.}
  \label{fig:RATjoint}
\end{figure*}

\section{Introduction}
\label{sec:intro}
Domain adaptation techniques such as fine-tuning \cite{freitag2016fast,luong-manning-2015-stanford} are highly effective at increasing in-domain performance of neural machine translation (NMT) systems, but are impractical in many realistic settings. For example, consider a single machine serving translations to thousands of customers, each with a private Translation Memory (TM). In this case, adapting, storing and loading large adapted models for each customer is computationally infeasible. In this paper we thus consider zero-shot adaptation instead, with a single general-domain model trained from heterogeneous sources that has access to the customer or domain specific TM only at inference time.

Our work builds on Retrieval Augmented Translation (RAT) \cite{RATsurvey,bulte-tezcan-2019-neural,xu-etal-2020-boosting,he-etal-2021-fast,cai-etal-2021-neural}, a paradigm which combines a translation model~\cite{NIPS2017_7181} with an external retriever module 
that retrieves the top-$k$ most similar source sentences from a TM (i.e. "fuzzy matches") ~\cite{farajian-etal-2017-multi,DBLP:journals/corr/GuWCL17,bulte-tezcan-2019-neural}. 
The encoder encodes the input sentence along with the translations of the top-$k$ fuzzy-matches and passes the resulting encodings to the decoder. 

Prior RAT methods for NMT have fallen into two camps: 
Early work \cite{bulte-tezcan-2019-neural, zhang-etal-2018-guiding}
concatenated the source sentence and the top-k fuzzy matches before encoding, 
relying on the encoder's self-attention to compare the source sentence to each target sentences and determine which target phrases are relevant for the translation. 
More recent work \cite{he-etal-2021-fast, cai-etal-2021-neural}
has opted to encode the source sentences and the top-k fuzzy matches 
\textit{independently}, effectively shifting the entire burden of determining which target phrases are relevant to the decoder.
We hypothesize that neither approach is ideal: 
In the first, the encoder has access to the information that we expect to be important (namely, the source and the fuzzy matches), but the self-attention also has potentially confusing/spurious connections. 
In the second, the encoder lacks the self-attention connections between the source and the fuzzy matches. 

To address these issues, we propose a novel architecture which has self-attention connections between the source sentence and each fuzzy-match, but not between fuzzy-matches. 
We denote this method \textbf{RAT with Selective Interactions} \textbf{(\ratsi)}. 
Our method is illustrated in \autoref{fig:RATjoint}, along with two previously discussed approaches.

Experiments in five English-German (En-De) domain-specific test sets \cite{aharoni2020unsupervised}
and 
seven English-French (En-Fr) domain specific test sets \cite{10.1162/tacl_a_00351},
for $k = \{3, 4, 5\}$,
demonstrate that our proposed method outperforms both prior approaches
in 32 out of 36 cases considered.
The proposed method outperforms the closest competitor by +0.82 to +1.75 BLEU for En-De and +1.57 to +1.93 for En-Fr.

\section{Method}

To isolate the effects of the underlying modeling strategy from the various tricks and implementation details employed in prior papers,
we build baseline models which distill the two primary modeling strategies used in prior works:

The first concatenates a source sentence with target-language fuzzy matches and then encodes the entire sequence,  as in \citet{bulte-tezcan-2019-neural} and
\citet{xu-etal-2020-boosting}.
In this approach, the cross-attention of the encoder must learn to find the relevant parts of target-language fuzzy-matches by comparing each fuzzy-match to the source sentence, while ignoring potential spurious fuzzy-match to fuzzy-match interactions (see the left diagram in \autoref{fig:RATjoint}).
We denote this method \textbf{\ratcat}.

The second encodes the source and each target-language fuzzy-match separately (with two distinct encoders),
and instead concatenates the encoded representations,
inspired by \citet{he-etal-2021-fast} and \citet{cai-etal-2021-neural}. In this approach, the spurious connections between the target-language fuzzy-matches are eliminated, but the connections between the source and each fuzzy-match are also eliminated, forcing the attention in the decoder to find the relevant portions in the fuzzy-match that are relevant to the source (see the center diagram in \autoref{fig:RATjoint}). We denote this method \textbf{\ratsep}.

Finally,
we propose a third method which attempts to build on the strengths of each of the prior methods. As in \ratsep, our method separately encodes (with the same encoder) the source and each target-language fuzzy-match; however, each fuzzy-match is jointly encoded with a copy of the source, as in \ratcat, allowing the encoder to find portions of the fuzzy-match which are relevant to the input. Finally, all the encoded inputs are concatenated and exposed to the decoder;
However, the encoding of the source is only provided to the encoder once, to avoid potentially spurious interactions between copies of the input (see the right diagram in \autoref{fig:RATjoint}).
We denote our proposed method \textbf{\ratsi}.

\section{Experimental Setup}

Our experiments are in two language directions: English-German (En-De) and English-French (En-Fr). 
We train models using the public WMT 2014~\cite{bojar-EtAl:2014:W14-33} data set, with $4.5M$ En-De sentences and $36M$ En-Fr sentences. 

During training, the model sees target-language fuzzy-match sentences from the same dataset it is being trained on (i.e. WMT14), but at inference, 
models must perform zero-shot adaptation to five En-De domain-specialized TMs\footnote{\texttt{Medical}, \texttt{Law}, \texttt{IT}, \texttt{Religion} and \texttt{Subtitles}.} and seven En-Fr domain-specialized TMs.\footnote{\texttt{News}, \texttt{Medical}, \texttt{Bank}, \texttt{Law}, \texttt{IT}, \texttt{TED} and \texttt{Religion}.} En-De data is taken from \newcite{aharoni2020unsupervised}, which is a re-split version of the multi-domain data set from \newcite{koehn-knowles-2017-six} while En-Fr data set is taken from the multi-domain data set of \newcite{10.1162/tacl_a_00351}.

To find target-language fuzzy matches for our model from domain specific TMs, 
we 
use Okapi BM25~\cite{robertson-bm25-2009}, a classical retrieval algorithm that performs search by computing lexical matches of the query with all sentences in the evidence, to obtain top-ranked sentences for each input.
To enable fast retrieval, we leverage the implementation provided by the ElasticSearch library.\footnote{\url{https://github.com/elastic/elasticsearch-py}}
Specifically, we built an index using source sentences of each TM,
and for every input source sentence, we collect top-$k$ similar source side sentences and then use their corresponding target side sentences as inputs to the model.

To explore how each method performs (and how robust they are) under different conditions, we run a full set of experiments for $k=\{3, 4, 5\}$. We train separate models for each language pair and $k$ value, and then apply that  model to each of the 5 (En-De) or 7 (En-Fr) domains.

We report translation quality with BLEU scores computed via \texttt{Sacrebleu}~\cite{post-2018-call}.\footnote{\texttt{nrefs:1|case:mixed|eff:no|tok:13a|\\smooth:exp|version:2.0.0}.}
We use \texttt{compare-mt} \cite{neubig-etal-2019-compare} to perform pairwise 
significance testing with \texttt{bootstrap = 1000} and
\texttt{prob\_thresh = 0.05} for all pairs.

All models employed transformers \cite{NIPS2017_7181} with 6 encoder and 6 decoder layers. Hidden size was set to  $1024$ and maximum input length truncated to $1024$ tokens. 
All models employed a joint source-target language subword vocabulary of size $32K$ using Sentencepiece algorithm~\cite{DBLP:conf/emnlp/KudoR18}.
Training details sufficient for replication are provided in Appendix \ref{appendix:parameters}.

\begin{table*}[tp]
\small
\centering
\begin{tabular}{l|llllllllllll}
Model       & k & IT    &  LAW  &   REL &  MED  &  SUBT & \multicolumn{1}{|c}{\textbf{Average}}\\\cline{0-7}
Baseline      &  n/a    & 27.92 & 35.59 & 11.26 & 30.74 & 19.46 & \multicolumn{1}{|c}{24.99}\\\hline
\ratcat  & \multirow{3}{*}{k=3}& \textbf{33.97} & 50.34 & 25.14 & 45.05 & 19.89 & \multicolumn{1}{|c}{34.88}\\
\ratsep     &                     & 32.78          & 49.04 & 22.92 & 44.28 & 20.48 & \multicolumn{1}{|c}{33.90}\\
\ratsi (this work)      &                     & 33.08 & \textbf{52.02*} & \textbf{26.40*} & \textbf{46.16} & \textbf{20.83*} & \multicolumn{1}{|c}{\textbf{35.70}}\\\cline{0-7}
\ratcat&\multirow{3}{*}{k=4}& {33.67}&49.59&23.40&44.87&\textbf{20.27}
&\multicolumn{1}{|c}{34.36}\\
\ratsep&&31.84&48.38&24.37&43.55&19.99&\multicolumn{1}{|c}{33.63}\\
\ratsi (this work)&& \textbf{33.68}&\textbf{52.00*}&\textbf{28.42*}&\textbf{46.13}&20.23&\multicolumn{1}{|c}{\textbf{36.09}}\\\cline{0-7}
\ratcat&\multirow{3}{*}{k=5}& 33.44&49.67&24.95&44.16&20.01
&\multicolumn{1}{|c}{34.45}\\
\ratsep&&30.84&47.92&23.91&44.10&20.27&\multicolumn{1}{|c}{33.41}\\
\ratsi (this work)&& \textbf{33.84}&\textbf{52.17*}&\textbf{27.53*}&\textbf{46.95*}&\textbf{20.49}&\multicolumn{1}{|c}{\textbf{36.20}}\\
\end{tabular}
\caption{BLEU scores for \textit{En-De} experiments. The best BLEU for RAT models with a specific top-k value is \textbf{bolded}, and "*" indicates the best result is statistically significant compared to both the other methods. The proposed method (\ratsi) produces the best results in 13/15 cases considered, with an average improvement of 1.43 BLEU over \ratcat and 2.35 BLEU over \ratsep.}\label{tab:app:FULL_indomain_wmta}
\end{table*}

\begin{table*}[tp]
\small
\centering
\begin{tabular}{l|llllllllllll}
Model & k &LAW & MED & IT & NEWS & BANK & REL & TED & \multicolumn{1}{|c}{\textbf{Average}}\\\hline
Baseline  & n/a &52.68&31.12&32.22&35.09&41.04&14.51&35.55&\multicolumn{1}{|c}{34.60}\\\hline
\ratcat&\multirow{3}{*}{k=3}& 66.32&37.09&39.91&35.09&49.01&61.83&36.39&\multicolumn{1}{|c}{46.52}\\
\ratsep&&64.93&37.06&38.02&\textbf{35.57}&49.14&53.34&37.29&\multicolumn{1}{|c}{45.05}\\
\ratsi (this work)&& \textbf{66.56}&\textbf{41.30*}&\textbf{40.61}&35.53&\textbf{50.19*}&\textbf{67.55*}&\textbf{37.42}&\multicolumn{1}{|c}{\textbf{48.45}}\\\hline
\ratcat&\multirow{3}{*}{k=4}& 65.71&37.31&38.71&34.60&49.43&63.55&36.10&\multicolumn{1}{|c}{46.49}\\
\ratsep&& 64.35&37.89&38.89&35.45&49.28&53.41&37.13&\multicolumn{1}{|c}{45.20}\\
\ratsi (this work)&& \textbf{66.63*}&\textbf{39.50*}&\textbf{41.90*}&{\textbf{35.71}}&\textbf{50.04}&\textbf{65.20*}&\textbf{37.47}&\multicolumn{1}{|c}{\textbf{48.06}}\\\hline
\ratcat&\multirow{3}{*}{k=5}& 65.60&37.35&38.74&34.46&49.33&63.21&36.08&\multicolumn{1}{|c}{46.40}\\
\ratsep&& 64.62&38.50&39.53&35.59&\textbf{49.97}&52.56&37.09&\multicolumn{1}{|c}{45.41}\\
\ratsi (this work)&& \textbf{67.03*}&\textbf{39.05}&\textbf{41.33*}&\textbf{35.93*}&{49.82}&\textbf{65.49*}&\textbf{37.90*}&\multicolumn{1}{|c}{\textbf{48.08}}\\
\end{tabular}
\caption{BLEU scores for En-Fr experiments. The best BLEU for RAT models with a specific top-k value is \textbf{bolded}, and "*" indicates the best result is statistically significant compared to both the other methods. The proposed method (\ratsi) produces the best results in 19/21 cases considered, with average improvements of 1.73 BLEU over \ratcat and 2.98 over \ratsep.}\label{tab:app:FULL_indomain_wmtaFrench}
\end{table*}

\section{Results}

Results for En-De are shown in Table~\ref{tab:app:FULL_indomain_wmta}  
and
results for En-Fr are shown in Table~\ref{tab:app:FULL_indomain_wmtaFrench}.

We observe several trends in the results. First,
our proposed \ratsi method outperforms both the \ratcat and \ratsep methods across 
both language pairs, having the best performance in 32/36 cases considered. 
In En-De, the proposed \ratsi method has an average improvement of 1.43 BLEU over \ratcat and 2.35 BLEU over \ratsep, while in 
En-Fr we observe an average improvement of 1.73 BLEU over \ratcat and 2.98 over \ratsep. These results support our hypothesis that attention connections between the source sentences and each fuzzy match are critical to translation quality and the connections between the fuzzy matches are actually harmful. 

Second, on average, $k=5$ produces the best results for the \ratsi method, but only by a small amount. However, considering individual language pair / domain combinations, there are many cases where $k=5$ does not produce the best results, sometimes by several BLEU points. We hypothesize that this is due to the different domains containing, on average, different amounts of relevant data. This observation underscores the importance of tuning $k$, as well as testing new RAT methods under a variety of conditions, including different $k$ values. 

Finally, consistent with prior work, we see large improvements for all online domain-adapted methods (\ratcat, \ratsep, and \ratsi) over the non-domain-adapted baseline, with improvements of up to $+13.85$ BLEU. This is not surprising, since the baseline model does not take advantage of any domain-specific data.

While not the focus of this work, we did a brief exploration of latency and observed only small differences in latency between \ratsi, \ratcat, and a baseline transformer (see Appendix~\ref{appendix:latency}).

\section{Related Work}

\citet{bulte-tezcan-2019-neural} proposed augmenting the input to NMT with target-language fuzzy-match sentences from a TM, concatenating the input and fuzzy-matches together. 
Their method was simpler than prior works such as \cite{zhang-etal-2018-guiding}, which manipulated n-gram probabilities based on their occurrence in the fuzzy-matches. 
\citet{xu-etal-2020-boosting} proposed several enhancements using the same architecture, including fine-tuning models and masking out or marking words not related to the input sentence, and matching arbitrarily large n-grams instead of sentences.

More recent work has explored using separate encoders for input and fuzzy-match \cite{he-etal-2021-fast,cai-etal-2021-neural}.
\citet{he-etal-2021-fast} also considers the realistic scenario where a TM may include noise, while \citet{cai-etal-2021-neural} explores finding target sentences in monolingual data instead of relying on a TM at inference time. 

\citet{Xia_Huang_Liu_Shi_2019} and \citet{xu-etal-2020-boosting} explore aspects of filtering fuzzy-matches by applying similarity thresholds, leveraging word alignment information \cite{zhang-etal-2018-guiding,xu-etal-2020-boosting,he-etal-2021-fast} or re-ranking with additional score (e.g. word overlapping) \cite{DBLP:conf/aaai/GuWCL18,zhang-etal-2018-guiding}.

Our work is related to the use of $k$-nearest-neighbor for NMT  \cite{khandelwal2021nearest,zheng-etal-2021-adaptive} but it is less expensive and does not require storage and search over a large data store of context representations and corresponding target tokens \cite{DBLP:journals/corr/abs-2105-14528}.

Other works have considered online adaptation outside the context of RAT, including \citet{vilar-2018-learning}, who proposes Learning Hidden Unit Contributions \cite{swietojanski2016learning} as a compact way to store many adaptations of the same general-domain model.
For an overview of fuzzy-match augmentation outside of NMT, we refer the reader to \citet{RATsurvey}.

TMs are often used in the localization industry to provide suggestions to translators in order to boost their productivity~\cite{federico-etal-2012-measuring}. Enhancing translation quality of  MT system by leveraging fuzzy-matches extracted from TMs has been explored widely for statistical MT~\cite{koehn-senellart-2010-convergence,mathur-etal-2013-online} and neural MT systems~\cite{farajian-etal-2017-multi,DBLP:journals/corr/GuWCL17,cao-xiong-2018-encoding,bulte-tezcan-2019-neural}.

\section{Conclusion}

Previous work in retrieval augmented translation
has used 
architectures 
which either have full connections between source and all fuzzy matches, 
or independently encode the source and each fuzzy match.
Based on our hypothesize that the attention connections between source and each fuzzy match are helpful, 
but that the the connections between different fuzzy matches are harmful,
we propose a new architecture (\ratsi) 
with the former connections but not the latter. 
Experiments on several language pairs, domains, and different numbers of fuzzy matches ($k$) 
demonstrate that \ratsi substantially outperforms the prior architectures.

\bibliography{main.bbl}
\bibliographystyle{acl_natbib}

\clearpage
\appendix

\addcontentsline{toc}{section}{Appendices}
\renewcommand{\thesubsection}{\Alph{subsection}}
\section*{Appendix}

\subsection{Model Parameters}\label{appendix:parameters}

We use the Adam optimizer~\cite{KingmaB14} with $\beta_1 = 0.9$, $\beta_2 = 0.98$ and $\epsilon = 10^{-9}$; and (ii) increase the learning rate linearly for the first $4K$ training steps and decrease it thereafter; (iii) use batch size of $32K$ source tokens and $32K$ target tokens. Checkpoints are saved after every $10K$ iterations during training. We train models with maximum of $300K$ iterations. 

We use dropout of $0.1$ and label-smoothing of $0.1$.
For evaluation we use Sacrebleu with hash value of \texttt{nrefs:1|case:mixed|eff:no|tok:13a|\\smooth:exp|version:2.0.0}.

\subsection{Latency}\label{appendix:latency}

While not the focus of this work, we did a preliminary study of latency, 
comparing 
a baseline transformer to \ratcat and \ratsi models during inference. 
We follow \newcite{domhan-etal-2020-sockeye} and measure latency values as the $90th$ percentile of inference time when translating each sentence individually (no batching). 
We run experiments on an \texttt{EC2} \texttt{p3.2xlarge} instance with a \texttt{Tesla V100 GPU} and report encoding latency results in Table~\ref{tab:inference_time}.
We use a batch size of 1 and k=3 for all experiments.

\begin{table}[ht]
\centering
\setlength{\tabcolsep}{3.5pt} %
\begin{tabular}{l|clllllll}
\hline
\multirow{1}{*}{Model} &Encoding Latency\\\hline
Transformer&14.80\\
\ratcat&15.23\\
\ratsi&17.48\\\hline
\end{tabular}
\caption{Encoding latency in milliseconds of models (lower is better).}\label{tab:inference_time}
\end{table}

\begin{table}[ht]
\centering
\setlength{\tabcolsep}{3.5pt} %
\begin{tabular}{l|llllllll}
\hline
\multirow{2}{*}{Detail Steps of \ratsi} &Encoding\\
&Latency\\\hline
Encode input and $k$ input-suggestion&14.91\\
Extra overheads&2.57\\\hline
Total time&17.48
\end{tabular}
\caption{Encoding latency of \ratsi in milliseconds (lower is better). We observe an amount of $14.7$\% latency on extra overheads including (1): Concatenate input and each of the suggestions into $k$ input-suggestion sequences (2): Extract $k$ suggestion encodings from $k$ input-suggestion encoding sequences and concatenate them with the encoding of the input.}\label{tab:ablation_encodingtime}
\end{table}

We observe a small increase of encoding latency by using \ratsi (i.e. $17.48$ ms) compared to of \ratcat.
However, the encoding time is a very small fraction of overall latency (see Table~\ref{tab:translation_time}) and thus this difference appears to be negligible.

\begin{table}
\centering
\setlength{\tabcolsep}{3.5pt} %
\begin{tabular}{l|llllllll}
\hline
\multirow{ 1}{*}{Model} &Translation Latency\\\hline
Transformer&574.02\\
\ratcat&597.28\\
\ratsi & 597.41\\\hline
\end{tabular}
\caption{Translation latency in milliseconds of \ratcat and our model \ratsi (lower is better). Batch size was set to one to simulate an on-demand system.}\label{tab:translation_time}
\end{table}

We find that \ratcat and \ratsi have nearly identical latencies, and each is only slightly slower than the baseline transformer (see Table~\ref{tab:translation_time}). 
This is somewhat surprising since both methods make the input to the decoder significantly longer. 
We hypothesize that we are under-utilizing the GPU in all cases, and thus the increased computations does not increase latency. Further investigation of this is left for future work.

\end{document}